\documentclass[lettersize,journal]{IEEEtran}

\IEEEoverridecommandlockouts                              

\usepackage{booktabs}    
\usepackage{amssymb} 

\usepackage{threeparttable} 

\usepackage{verbatim}
\usepackage{graphicx} 
\usepackage{subfigure} 
\usepackage{float}
\graphicspath{{figures}}
\usepackage{multirow}

\usepackage[linesnumbered,ruled,vlined]{algorithm2e}
\usepackage{algpseudocode}
\usepackage{amsmath}
\usepackage{xcolor}
\usepackage{setspace}

\usepackage{biblatex}
\usepackage[colorlinks,linkcolor=black,anchorcolor=blue,citecolor=black]{hyperref}

\usepackage{lettrine}

\def\BibTeX{{\rm B\kern-.05em{\sc i\kern-.025em b}\kern-.08em
		T\kern-.1667em\lower.7ex\hbox{E}\kern-.125emX}}
\newcommand\figcaption{\def\@captype{figure}\caption}
\newcommand\tabcaption{\def\@captype{table}\caption}

\title{
MotionBEV: Attention-Aware Online LiDAR Moving Object Segmentation with Bird's Eye View based Appearance and Motion Features
}

\author{Bo Zhou$^{*}$, Jiapeng Xie$^{*}$, Yan Pan, Jiajie Wu, and Chuanzhao Lu
\thanks{This work is partly supported by the National Natural Science Foundation (NNSF) of China under the Grants No. 62073075. ($^{*}$Bo Zhou and Jiapeng Xie contributed equally to this work.) (Corresponding author: Bo Zhou.)}
\thanks{All the authors are with the School of Automation, Southeast University, Nanjing 210096, P. R. China (emails: zhoubo@seu.edu.cn; xiejiapeng@seu.edu.cn; yanpan@seu.edu.cn; jiajiewu@seu.edu.cn; lucz@seu.edu.cn)}%
}

\UseRawInputEncoding
\begin{document}

\maketitle
\thispagestyle{empty}
\pagestyle{empty}

\begin{abstract}

Identifying moving objects is an essential capability for autonomous systems, as it provides critical information for pose estimation, navigation, collision avoidance, and static map construction. In this paper, we present MotionBEV, a fast and accurate framework for LiDAR moving object segmentation, which segments moving objects with appearance and motion features in the bird's eye view (BEV) domain. Our approach converts 3D LiDAR scans into a 2D polar BEV representation to improve computational efficiency. Specifically, we learn appearance features with a simplified PointNet and compute motion features through the height differences of consecutive frames of point clouds projected onto vertical columns in the polar BEV coordinate system. We employ a dual-branch network bridged by the Appearance-Motion Co-attention Module (AMCM) to adaptively fuse the spatio-temporal information from appearance and motion features. Our approach achieves state-of-the-art performance on the SemanticKITTI-MOS benchmark. Furthermore, to demonstrate the practical effectiveness of our method, we provide a LiDAR-MOS dataset recorded by a solid-state LiDAR, which features non-repetitive scanning patterns and a small field of view. 

\end{abstract}


\section{Introduction}

\lettrine[lines=2]{D}{ynamic} objects such as pedestrians, cyclists, and moving vehicles are frequently present in traffic scenes. They can cause errors in localization and mapping \cite{zhang2014loam,chen2019suma++}, and hinder downstream tasks like obstacle avoidance \cite{guo2022obstacle} and path planning \cite{luo2018porca}. In this context, online moving object segmentation (MOS) plays a critical role in enabling autonomous systems to obtain a more accurate perception of the environment and make reliable decisions.

This paper focuses on the task of LiDAR moving object segmentation (LiDAR-MOS), which involves identifying objects in a scene that are currently in motion using LiDAR (Light Detection and Ranging) sensors. This task is challenging due to the sparsity, uncertain distribution, and the noise of LiDAR point cloud, as well as the complexity and diversity of dynamic scenes. To fully understand the dynamics of the environment, it is necessary to exploit 4D spatio-temporal information from sequential LiDAR frames.

Some map cleaning methods \cite{kim2020remove, lim2021erasor} reject moving points with geometry information, but they often rely on global maps and can only run offline. Recent approaches leverage learning-based methods for map-free moving object segmentation. Some point-based methods \cite{mersch2022receding, sun2020pointmoseg} extract features directly from the sequence of point clouds, but these methods typically require a significant amount of computational resources and are not suitable for real-time applications. Chen \emph{et al.} \cite{chen2021moving} proposed LMNet, a method that leverages residual range images to capture temporal information and aligns features through direct concatenation. However, range view (RV) based methods may suffer from boundary-blurring issues resulting from back-projection, and the motion cues derived from residual range images may not be optimal for representing temporal information due to their sensitivity to changes in distance. Moreover, the direct feature alignment of multi-modality features can be limited in accuracy due to redundant and invalid information in motion features. Some subsequent studies \cite{sun2022efficient,kim2022rvmos} have attempted to enhance the fusion of spatial-temporal information by introducing attention mechanisms and incorporating semantic information. However, these approaches still heavily rely on the quality of the motion features, resulting in decreased accuracy when motion features are of poor quality. 

Motivated by the aforementioned challenges, we propose a Bird's Eye View (BEV) based moving object segmentation method called MotionBEV. We convert 3D LiDAR point clouds into a 2D BEV representation to improve computational efficiency. The BEV representation maintains a consistent object size regardless of distance and has better spatial consistency compared to the RV representation. Such a view can provide a rich spatial context that is easy to interpret and process.

The contributions of this paper are summarized as follows:
\begin{itemize}
\item [$\bullet$]
We propose a bird's eye view-based method that exploits high-quality spatio-temporal information for LiDAR-MOS from appearance and motion features. Specifically, the method learns appearance features for each grid cell in the polar BEV images using a simplified PointNet \cite{qi2017pointnet}, while extracting motion features through the height differences of vertical columns. Temporal information captured from such BEV-based motion features is robust to distance changes. 
\item [$\bullet$]
We design a dual-branch network bridged by the Appearance-Motion Co-attention Module (AMCM) to adaptively fuse appearance and motion features. To avoid excessive reliance on motion features, the AMCM dynamically assigns importance weights to appearance and motion features to balance their contributions. Furthermore, the AMCM enhances appearance features using motion features in an attention mechanism, ensuring that the fusion of appearance and motion features is effective and mutually reinforcing.
\item [$\bullet$]
The proposed method outperforms existing baselines on the SemanticKITTI-MOS benchmark \cite{chen2021moving, behley2019semantickitti}, with 69.7\% IoU for moving class and an average inference time of 23ms (on an RTX 3090 GPU). In addition, we evaluate our method on a dataset collected by a solid-state LiDAR and demonstrate its practical effectiveness on LiDARs with non-repetitive scanning patterns and small filed of view.
\end{itemize}

Our code will be publicly available \footnote{\href{https://github.com/xiekkki/motionbev}{https://github.com/xiekkki/motionbev}}.


\section{Related Works}

\begin{figure*}[t]  
	\centering
	\includegraphics[width=1\linewidth]{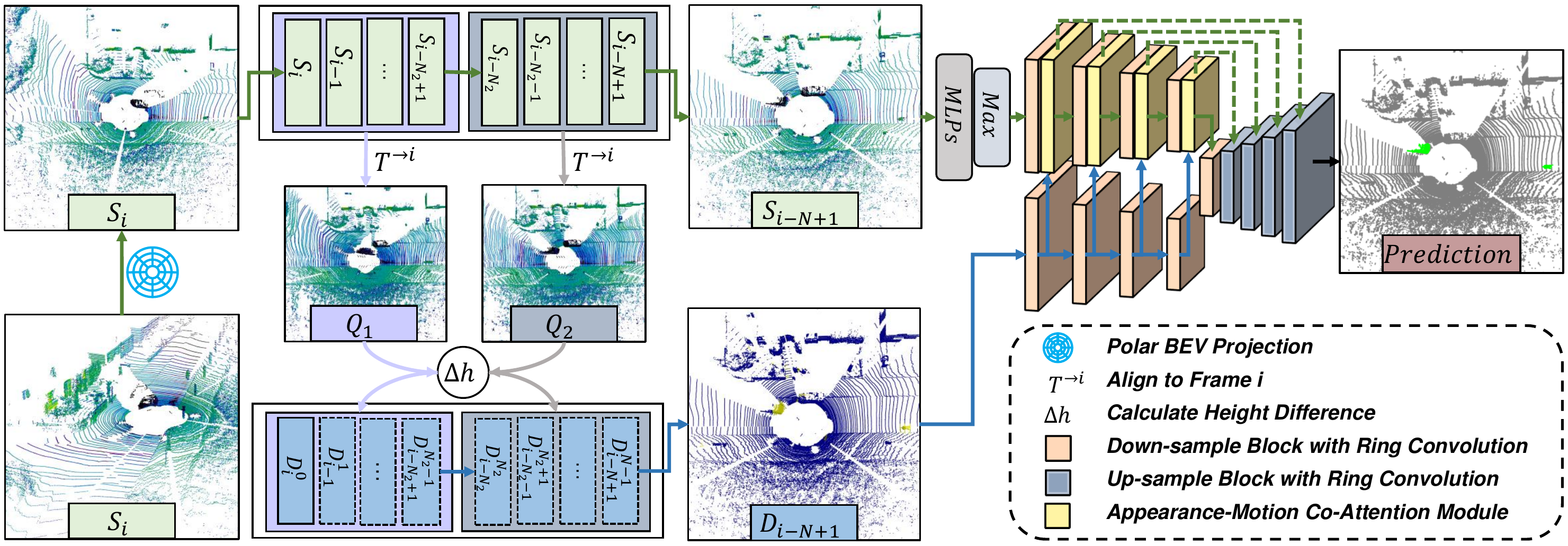}
	\vspace{-20pt}
	\caption{Overview of MotionBEV. For each LiDAR scan, we first partition the points into grids using the polar BEV coordinates. We push the scan into the temporal window and transform all scans of the window to the current viewpoint. After $N$ iterations of movement, we obtain the motion features with $N$ channels. We then pop the scan out and employ PointNet to learn the appearance features. The appearance and motion features are fed into a dual-branch network bridged by AMCM to explore spatio-temporal information. Finally, the CNN outputs a quantized prediction and we decode it to the point domain.}
	\label{Fig: pipeline}	
	\vspace{-15pt}
\end{figure*}
While significant progress has been made in the field of image and video-based moving object segmentation \cite{yang2021learning,li2019motion,giraldo2020graph,zhao2022modeling}, challenges still remain for LiDAR-MOS task due to the sparse and uneven distribution of point clouds. This chapter focuses on LiDAR-based moving object segmentation, which can be broadly categorized into two main approaches: geometry-based methods and learning-based methods.

\vspace{-10pt}
\subsection{Geometry-based Methods}

Geometry-based methods do not require training data but are typically offline and rely on pre-built maps. Lim \emph{et al.} \cite{lim2021erasor} removes dynamic objects by checking the pseudo occupancy ratio of each sector in the LiDAR scan and recovers the ground plane through region growing. Schauer and N{\"u}chter \cite{schauer2018peopleremover} propose a ray casting-based method that estimates the occupancy probability in the grid space to distinguish static and dynamic points. To improve computational efficiency, the visibility-based method \cite{pomerleau2014long} projects the point cloud onto a range image, and clears dynamic points by comparing the visibility difference between frames and maps. Kim \emph{et al.} \cite{kim2020remove} propose a remove-then-revert mechanism that iteratively retains static points from mistakenly removed points using a multi-resolution range image. Fan \emph{et al.} \cite{fan2022dynamicfilter} fuse ray casting-based and visibility-based methods to achieve high-precision dynamic object segmentation. Chen \emph{et al.} \cite{chen2022automatic} propose an automatic annotation method, using clustering-based object segmentation and tracking-based association to provide training labels for learning-based MOS. In general, these methods are ideal for static map construction but are not suitable for online MOS.

\vspace{-10pt}
\subsection{Learning-based Methods}

Learning-based methods learn effective information directly from data with trainable deep neural networks, enabling the detection of dynamic objects without pre-built maps.

LiDAR semantic segmentation \cite{zhang2020polarnet,cortinhal2020salsanext,hu2020randla,zhu2021cylindrical} is a closely related task to LiDAR-MOS. However, most semantic segmentation methods can only distinguish between movable and immovable objects, such as vehicles and buildings, but cannot differentiate between moving and non-moving objects, such as moving cars and parked cars. Scene-flow methods \cite{ding2022self,li2022rigidflow,dong2022exploiting} estimate 3D scene flow between consecutive point clouds, enabling the identification of moving points. However, most scene-flow methods consider only two subsequent frames, resulting in limited accuracy for slowly moving objects. Some point-based methods \cite{mersch2022receding, sun2020pointmoseg, kreutz2023unsupervised, wang2023insmos} extract temporal information directly from sequences of point clouds to improve accuracy and generalization. However, due to the high dimensionality of point clouds, these methods often suffer from computational inefficiencies and limited real-time performance.

Recently, Chen \emph{et al.} \cite{chen2021moving} release a moving object segmentation dataset and benchmark based on SemanticKITTI \cite{behley2019semantickitti}. At the same time, they propose LMNet, which extracts temporal features from the residual range images for online MOS. To better exploit spatio-temporal information, Sun \emph{et al.} \cite{sun2022efficient} presents a dual-branch structure to separately process spatial and temporal information and fuse them with motion-guided attention modules. Kim \emph{et al.} \cite{kim2022rvmos} design a network that incorporates semantic information to further improve the MOS performance. These methods all exploit spatio-temporal information of range images, which can suffer from boundary-blurring issues during the back-projection process and be sensitive to changes in distance. 

Compare to the range view projection, bird's eye view (BEV) projection provides a more intuitive representation of object motion and spatial relationships in the scene. 
Mohapatra \emph{et al.} \cite{mohapatra2021limoseg} introduced a BEV-based method that exploits the disparity between two successive frames with a residual computation layer. Although this method runs very fast, it exhibits relatively low accuracy compared to RV-based methods. To improve the MOS performance, our approach leverages grid height differences between BEV images to extract effective temporal information and uses a lightweight PointNet \cite{qi2017pointnet} to learn appearance features for each vertical grid. The spatial and temporal features are deeply fused with the appearance-motion co-attention module. To handle complex scene changes, we add co-attention gates \cite{yang2021learning} before the motion-guided attention modules \cite{li2019motion} to suppress the interference from redundant and invalid information. Benefitting from the BEV representation, our method achieves state-of-the-art performance without any post-processing, while also offering faster computation compared to range view-based methods.


\vspace{-5pt}
\section{Proposed Approach}

In this section, we introduce our bird's eye view-based framework for moving object segmentation, whose overall pipeline is illustrated in Fig. \ref{Fig: pipeline}. First, in Section III-A, we describe the projection of 3D LiDAR point clouds to the polar bird's eye view image. Then, we extract temporal information by computing the height differences of the projected local maps in Section III-B. The structure and modules of our network are presented in Section III-C, and the details of our network training are described in Section III-D.

\subsection{Input Representation} 
Given the LiDAR scan of the $i^{th}$ frame in the point cloud sequence, denoted as $S_i = {\{ p_j \in \mathbb R^4 \}}_{j=0}^{M-1},p_j=[x_j,y_j,z_j,1]^T$, which contains $M$ points represented in homogeneous coordinates, and the past $N-1$ consecutive frames $S_{i-1},S_{i-2},...,S_{i-N+1}$, the goal of MOS is to determine which points in $S_i$ are actually moving. The current frame $S_i$ provides appearance features for prediction, while the consecutive $N$ frames $S_{i},S_{i-1},...,S_{i-N+1}$ provide motion features. First, we align the past frames to the viewpoint of $S_i$ using the relative pose transformations $T \in \mathbb R^{4 \times 4}$ estimated by LiDAR odometry as:
\begin{equation}\label{eq-1} 
	\begin{aligned}
		S^{\to i}_{i-n} &= {\{ T^{i}_{i-n} p_j\  | \ p_j \in S_{i-n} \}},\ \mathrm{where}\\ 
		T^{i}_{i-n} &= \prod_{k=1}^{n} T^{i-k+1}_{i-n},n \in {\{ 1, 2, ...,N-1 \}}
	\end{aligned}
\end{equation}

To achieve online MOS, we project the 3D LiDAR points onto 2D image coordinates. As suggested in \cite{zhang2020polarnet}, we adopt the polar bird's eye view for point cloud coordinate partitioning to balance the uneven distribution of LiDAR points in space. For each point $p_j = (x_j,y_j,z_j) \in S_i$ represented in Cartesian coordinates, we use the following equations to convert it into a representation in polar coordinates:
\begin{equation}\label{eq-2}
\begin{pmatrix} \rho_j \\ \theta_j\\ z_j \end{pmatrix} = \begin{pmatrix} \sqrt{x^2_j + y^2_j} \\ \arctan{(y_j, x_j)}\\ z_j \end{pmatrix}
\end{equation}

Each point $p_j$ is assigned to the corresponding grid of the polar BEV image according to its polar coordinates:
\begin{equation}\label{eq-3}
	\begin{aligned}
		S_{(u,v),i}&=\{ p_j \ | \   p_j \in S_i, \\
		{\rho_{max}-\rho_{min} \over w}\cdot (u-1) & \leqslant \rho_j < {\rho_{max}-\rho_{min} \over w}\cdot u, \\
		{\theta_{max}-\theta_{min} \over h}\cdot (v-1) & \leqslant \theta_j< {\theta_{max}-\theta_{min} \over h}\cdot v\}
	\end{aligned}
\end{equation}
where $S_{(u,v),i}$ denotes all points of $S_i$ contained in the $(u, v)^{th}$ grid, $(h, w)$ are the height and width of the polar BEV image, $(\theta_{max},\theta_{min})$ are the maximum and minimum limits of the angle, and $(\rho_{max},\rho_{min})$ are the maximum and minimum values of the distance. In the experiment section, we will study the impact of the size of the BEV image on MOS performance. The limits of angle and distance can be specified based on the characteristics of the LiDAR sensor and environment. 

\vspace{-10pt}
\subsection{Motion Features Generation}
The generation of motion features aims to extract temporal information from consecutive LiDAR frames to provide clues for moving object segmentation. LMNet \cite{chen2021moving} achieves this by calculating residual range images, but we found that these images are not efficient in representing temporal information due to their sensitivity to changes in distance. To address this issue, we propose a method that involves calculating the height difference between bird's eye view images to generate motion features.

To mitigate the effects of sparse point clouds, we refrain from directly comparing differences between frames. Instead, we maintain two adjacent temporal windows, $Q_1$ and $Q_2$, with equal lengths of $N_2=N/2$, and generate motion features by examining the height differences of corresponding grid cells in $Q_1$ and $Q_2$. We first compensate for ego-motion by relative pose transformation and align $Q_1$ and $Q_2$ to the current local coordinate system. The polar BEV images are calculated according to the partition results of Eq. (\ref{eq-2}) and Eq. (\ref{eq-3}), where each pixel value $I_{(u,v),i}$ represents the height occupied by the $(u,v)_{th}$ grid in $Q_i$:
\begin{equation}\label{eq-4}
	\begin{aligned}
		I_{(u,v), i} = &\mathrm{Max}\{Z_{(u,v),i}\}-\mathrm{Min}\{Z_{(u,v),i}\},\ \mathrm{where}\\ 
		Z_{(u,v), i} = \{& z_j \in p_j \ | \ p_j \in  Q_{(u,v),i}, z_{min} < z_j < z_{max}\}
	\end{aligned}
\end{equation}

We obtain the region of interest by limiting the range of $z$ to the interval $(z_{min},z_{max})$, which is the specific location where moving objects like vehicles and pedestrians usually appear. For the SemanticKITTI dataset, we take $(z_{min},z_{max})=(-4,2)$. We subtract the projected BEV images $I_1$ and $I_2$ to obtain the residuals:
\begin{equation}\label{eq-4-11}
	\small
	\begin{aligned}
		D_{(u,v), i}^{0}, D_{(u,v), i-1}^{1}, ..., D_{(u,v), i-N_2+1}^{N_2-1} &= I_{(u,v), 1} - I_{(u,v), 2}, &\\
		D_{(u,v), i-N_2}^{N_2}, D_{(u,v), i-N_2-1}^{N_2+1}, ..., D_{(u,v), i-N+1}^{N-1} &= I_{(u,v), 2} - I_{(u,v), 1}&
	\end{aligned}
\end{equation}
where $D_{(u,v), i}^{k}$ represents the motion feature in the $(u, v)^{th}$ grid of the $k^{th}$ channel for the $i^{th}$ frame. The value of $k$ is determined based on the frame's position in the temporal window. We shift both temporal windows to the next position as each new frame arrives. Each frame stays within the temporal window for $N$ cycles, thus ensuring that the motion features retain a sequential characteristic with a length of $N$. To reduce the interference of noise and occlusion, we discard regions with residual values that are too large ($D_{(u,v), i} > $ 4) or too small ($D_{(u,v), i} < $ 0.4), as well as regions that might be blocked by nearby objects and therefore have too few points (less than 5). 

It is worth noting that generating complete motion features requires the scan to pass through a temporal window of length N, which introduces a fixed delay of $(N-1) \times T$, where $T$ is the measurement period of the sensor. To offer a delay-free solution, one can opt to utilize only the motion features from the first channel, as $D_{i}^{0}$ can be computed immediately when $S_{i}$ enters the temporal window. Nevertheless, adopting this approach may lead to a trade-off in segmentation accuracy.

\begin{figure}[t]  
	\centering
	\includegraphics[width=1\linewidth]{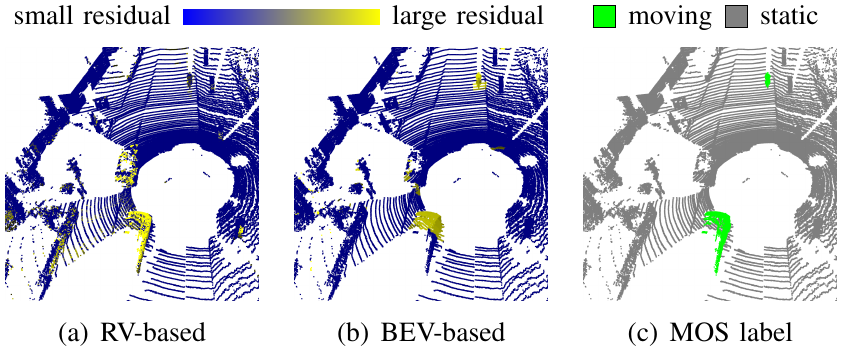}
	\vspace{-20pt}
	\caption{Visualization and comparison of range view-based motion features (RV-based) and bird's eye view-based motion features (BEV-based).}
	\label{Fig: residual}	
	\vspace{-15pt}
\end{figure}

For comparison, we apply back-projection to both the range view-based motion features and our bird's eye view-based motion features to 3D space, as shown in Fig. \ref{Fig: residual}. Our method attributes more salient motion features to moving cars and bicycles, whereas the range view-based method tends to assign larger residuals to the nearby parked car compared to the bicycle driving far away, making it difficult to distinguish whether distant objects are stationary or moving. Despite the distinctiveness of our motion features for moving object segmentation, noise and occlusion could result in erroneous residuals, mainly for points hidden behind objects or accidentally appearing. Therefore, it is not recommended to use motion features as MOS results directly. Nevertheless, most of these erroneous residuals can be distinguished by the subsequent CNN network with the aid of appearance features as they tend to appear on immovable objects.

\vspace{-10pt}
\subsection{Network Structure}

Our network structure is built upon PolarNet \cite{zhang2020polarnet}, a polar bird's eye view-based method which achieves end-to-end LiDAR semantic segmentation through an encoder-decoder architecture. In order to explore spatio-temporal information for MOS, we modify PolarNet into a dual-branch structure and adaptively fuse appearance and motion features with the appearance-motion co-attention module.

\subsubsection{Appearance Features Encoding}
In contrast to the manual extraction of appearance features, we adopt a simplified PointNet \cite{qi2017pointnet} followed by max-pooling to learn the point cloud distribution across the vertical column of a grid. For the grid cell of $(u,v)$ coordinate in the polar BEV image, the appearance features can be represented as follows:
\begin{equation}\label{eq-5}
	F^a_{(u,v),i} = \mathrm{MaxPool} \{ \mathrm{MLPs}(p_j) \ | \  p_j \in S_{(u,v),i}  \}
\end{equation}

Since the representation is learned in the polar coordinate system, the left and right ends of the bird's eye view image should be connected in physical space. We use the same ring convolution kernel as \cite{zhang2020polarnet} to extract complete spatial information and replace all the conventional convolution kernels in the network.

\begin{figure}[t]  
	\centering
	\includegraphics[width=1\linewidth]{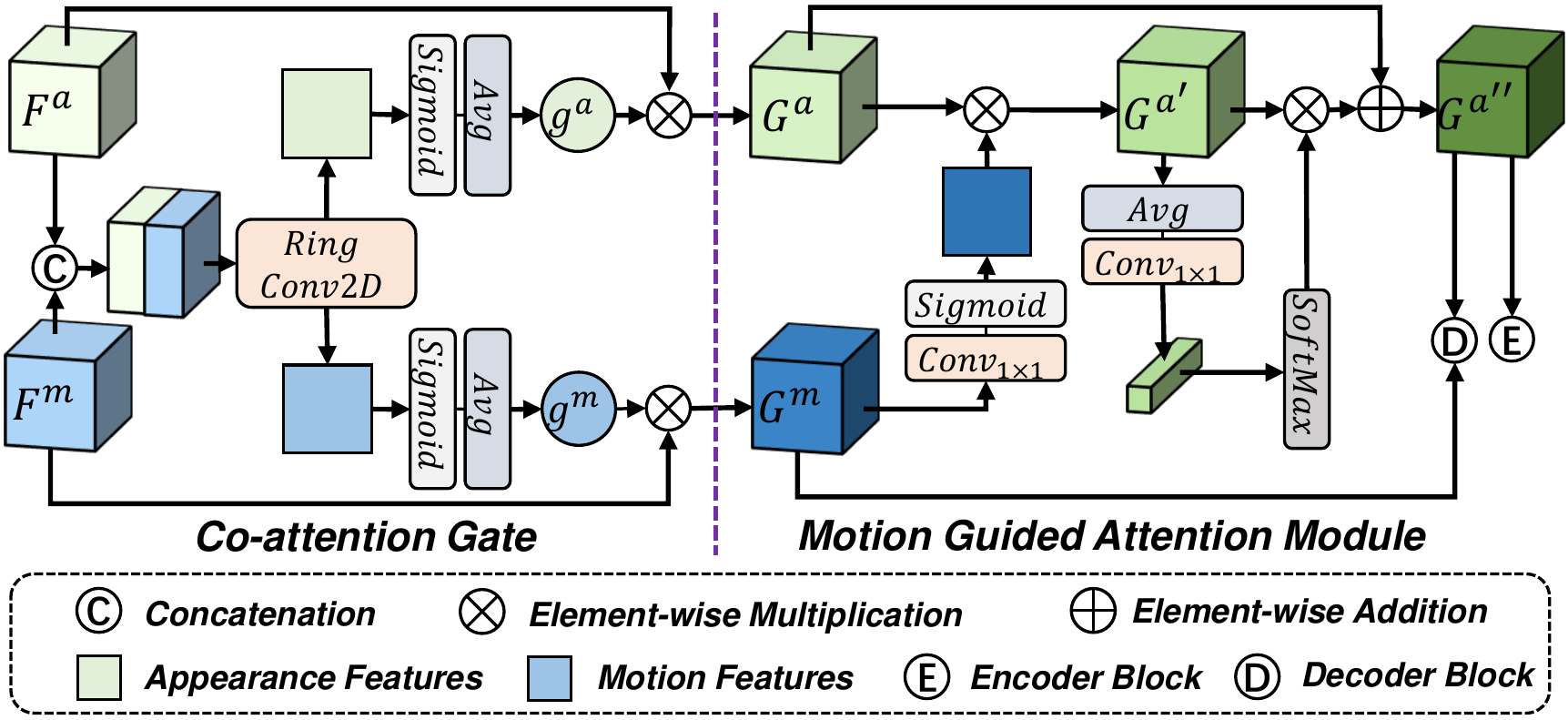}
	\vspace{-20pt}
	\caption{Structure of the Appearance-Motion Co-attention Module (AMCM).}
	\label{Fig: amcm}	
	\vspace{-15pt}
\end{figure}

\subsubsection{Appearance-Motion Co-attention Module}
We utilize the Appearance-Motion Co-attention Module (AMCM) to facilitate interaction between motion-appearance features from two different branches. Inspired by visual segmentation methods, our AMCM consists of two parts: the co-attention gate \cite{yang2021learning} and the motion guided attention module \cite{li2019motion}, as shown in Fig. \ref{Fig: amcm}.

First, the co-attention gate is used to adaptively compute the importance of appearance features and motion features, in order to balance their contributions and suppress the redundant and misleading information. Given the appearance features $F_i^a$ and motion features $F_i^m$ from the $i^{th}$ layer, we capture the relative relationship between multi-modal features using cross-channel concatenation and ring convolution operations, resulting in a fused feature $H \in \mathbb R^{h \times w \times 2}$. Next, we split feature $H$ into two sub-branches, and apply the sigmoid function and global average pooling on each channel to obtain a pair of co-attention scores $g_i^a$ and $g_i^m$, reflecting the importance of each modality feature. We combine the appearance features and motion features to construct the gating function:
\begin{equation}\label{eq-6}
g_i = \mathrm{Avg}(\mathrm{Sigmoid}(\mathrm{Conv}(\mathrm{Cat}(F_i^a, F_i^m))))
\end{equation}

where $g_i$ contains the pair of co-attention scores $g_i^a$ and $g_i^m$. Higher co-attention scores indicate that the corresponding modality feature contains more effective information for accurate segmentation. We apply the co-attention scores to the features, generating gated appearance features $G^a_i$ and gated motion features $G^m_i$, where
\begin{equation}\label{eq-7}
G^a_i = F_i^a \otimes g_i^a, G^m_i = F_i^m \otimes g_i^m
\end{equation}

The motion guided attention module follows the co-attention gate, and utilizes motion features to emphasize important positions or elements in the appearance features. Specifically, we first fuse the gated motion features $G^m_i$ and the gated appearance features $G^a_i$ by spatial attention, generating motion salient feature ${G_i^a}^{\prime}$:
\begin{equation}\label{eq-8}
{G_i^a}^{\prime} = G^a_i \otimes \mathrm{Sigmoid}(\mathrm{Conv_{1\times1}}(G^m_i))
\end{equation}

Then, we use channel attention to enhance the response of key attributes, generating the final spatio-temporal fused feature ${G_i^a}^{\prime \prime}$ of size $C \times h \times w$, with the following formula:
\begin{equation}\label{eq-9}
{G_i^a}^{\prime \prime} = {G_i^a}^{\prime} \otimes [\mathrm{Softmax}(\mathrm{Conv_{1\times1}}(\mathrm{Avg}({G_i^a}^{\prime}))) \cdot C] + G^a_i
\end{equation}

By combining the co-attention gate and motion guided attention module, the AMCM is able to perform adaptive multi-modality features fusion.

\vspace{-10pt}
\subsection{Implementation Details}
We use PyTorch to build our network and train it on an NVIDIA RTX 3090 GPU with a batch size of 8. The input sequences for training are randomly shuffled before each epoch. Widely-used data augmentation techniques such as random flipping, rotation, and small-scale translation are applied to enhance the training data. The loss function of the network is a linear combination of weighted cross-entropy ($\mathcal L_{wce}$) and Lov{\'a}sz-Softmax ($\mathcal L_{ls}$) \cite{berman2018lovasz} losses:
\begin{equation}\label{eq-10}
\mathcal L = \mathcal L_{wce} + \mathcal L_{ls},
\end{equation}
where the weighted cross-entropy loss is defined as:
\begin{equation}\label{eq-10-1}
	\mathcal L_{wce}(y, \hat y) = - \sum \alpha_{i}p(y_i)\mathrm{log}(p(\hat y_i)), \alpha_i = 1 / \sqrt{f_i},
\end{equation}
and the Lov{\'a}sz-Softmax loss is given by:
\begin{equation}\label{eq-10-2}
	\mathcal L_{ls} = {1 \over |C|} \sum_{c \in C} \bar{\Delta_{J_c}} (m(c)), m_i(c)=
	\left\{
		\begin{array}{lr}
			1-x_i(c) \   \mathrm{if\ } c=y_i(c)\\
			x_i(c)  \ \ \ \ \ \    \mathrm{otherwise}
		\end{array} 
	\right. 
\end{equation}
In the equations, $y_i$ and $\hat{y_i}$ represent the true and predicted labels, and $f_i$ is the frequency of the $i^{th}$ class. $|C|$ is the number of classes, $\bar{\Delta_{J_c}}$ represents the Lov{\'a}sz extension of the Jaccard index. Additionally, $x_i(c) \in [0, 1]$ and $yi(c) \in \{−1, 1\}$  represent the predicted probability and ground truth label of pixel $i$ for class $c$, respectively.

We use stochastic gradient descent (SGD) to minimize $\mathcal L_{wce}$ and $\mathcal L_{ls}$, with a momentum of 0.9 and weight decay of 0.0001. The initial learning rate is set to 0.005, and it is decayed by a factor of 0.99 after each epoch. No pre-trained weights are used, and the network is trained from scratch until the validation loss converges.


\section{Experiments}

This section presents the experimental evaluation of our method for moving object segmentation. First, we introduce the datasets and evaluation metrics in Section IV-A. In Section IV-B, we provide quantitative and qualitative comparisons between our method and other state-of-the-art methods. In Section IV-C, we conduct an ablation study to demonstrate the effectiveness of our BEV-based motion features generation method, as well as the contributions of different components in the network to the overall performance. In Section IV-D, we show the practical effect of our method on a solid-state LiDAR. Lastly, in Section IV-E, we showcase the runtime performance of our method.

\vspace{-5pt}
\subsection{Experiment Setups}

\textbf{Datasets-SemanticKITTI.} We evaluate our method by comparing its performance with state-of-the-art methods on the SemanticKITTI-MOS \cite{chen2021moving} dataset.  The original SemanticKITTI dataset \cite{behley2019semantickitti,geiger2012cvpr} comprises 22 labeled point cloud sequences collected by a single Velodyne HDL-64E LiDAR. SemanticKITTI-MOS maps all semantic classes to two categories: moving and static. The dataset is split into train sequences 00-07, 09-10 (19,130 frames), validation sequence 08 (4,071 frames), and test sequences 11-21 (20,351 frames). The odometry estimation is obtained from Suma \cite{behley2018efficient}.

\textbf{Datasets-SipailouCampus.} To further assess the effectiveness of our method across different LiDAR sensors, we collected a dataset with a solid-state LiDAR Livox Avia mounted on an unmanned vehicle, at different areas of Southeast University Sipailou Campus. The Livox Avia has a narrower field of view (70.4$^{\circ}$ horizontally and 77.2$^{\circ}$ vertically) and is equipped with non-repetitive scanning mode, which presents new challenges for MOS. We manually annotate all sequences with moving labels, resulting in a dataset of 26,279 frames, of which 15,585 frames contain dynamic objects, and each frame has 24,000 points. We use 5 sequences (16,887 frames) for training, 1 sequence (3,191 frames) for validation, and 2 sequences (6,201 frames) for testing. The odometry estimation is obtained from FAST-LIO \cite{xu2022fast}. We will release this dataset to facilitate further research.

\textbf{Evaluation Metrics.} Following the previous work \cite{chen2021moving}, we use the Jaccard Index or Intersection-over-Union (IoU) metric \cite{everingham2010pascal} over moving objects to quantify MOS performance:

\begin{equation}\label{eq-11}
\mathrm{IoU= {TP \over{TP+FP+FN}}}
\end{equation}
where TP, FP, and FN represent the number of true positives, false positives, and false negatives in the prediction of the moving class, respectively.

\vspace{-5pt}
\subsection{Evaluation and Comparisons on SemanticKITTI}
\begin{figure*}[t]  
	\centering
	\includegraphics[width=1\linewidth]{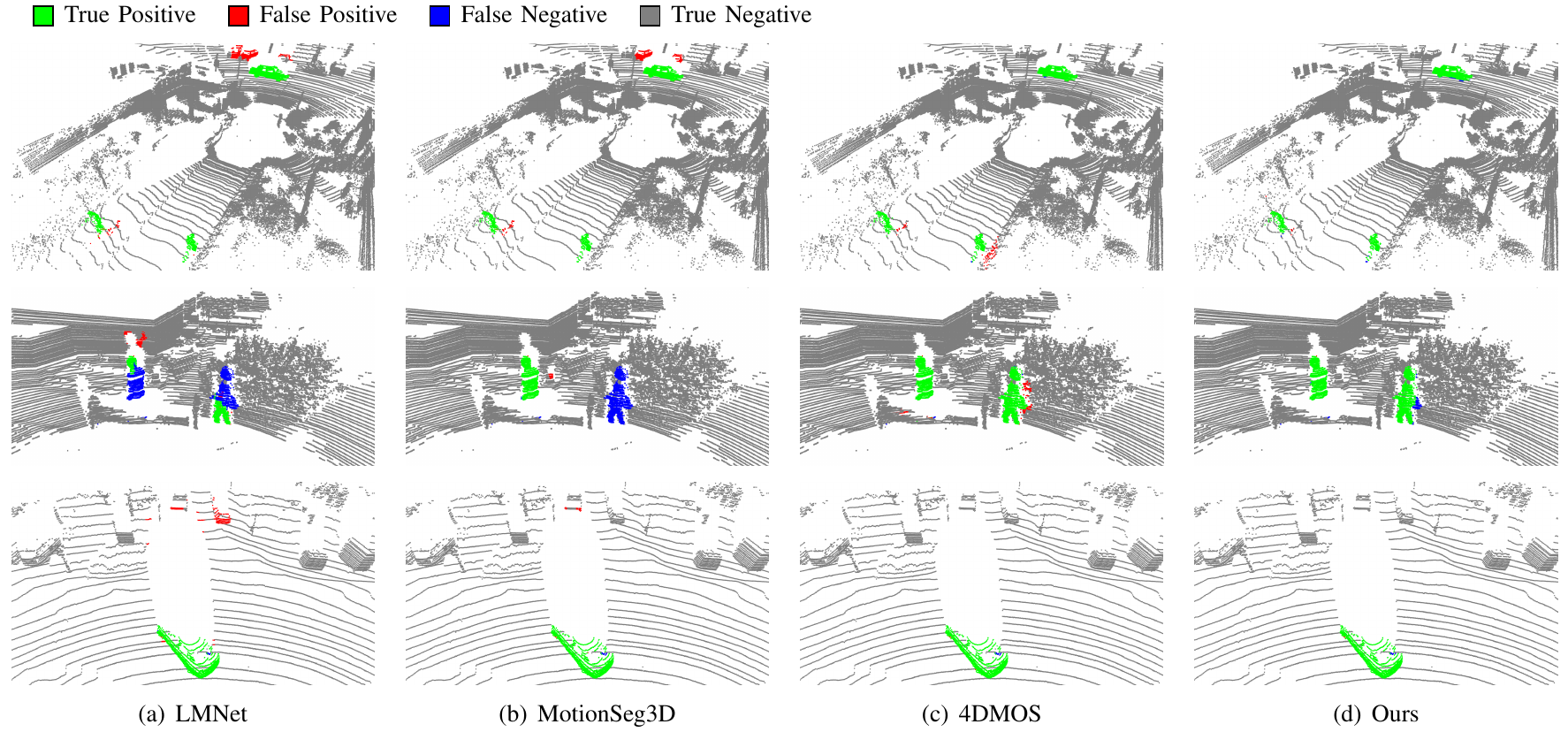}
	\vspace{-20pt}
	\caption{Qualitative results of different methods for LiDAR-MOS on SemanticKITTI validation set. Best viewed in color.}
	\label{Fig: comparison}	
	\vspace{-15pt}
\end{figure*}

We evaluate our method by submitting moving object segmentation predictions on the SemanticKITTI-MOS benchmark. We consider two alternatives for our method: one utilizes only $D_{i}^{0}$ (referred to as "without delay"), while the other incorporates complete motion features with a temporal window size of 8. Our method is compared against various baseline methods, including a) range view-based methods: LMNet \cite{chen2021moving}, MotionSeg3D v1 (with kNN) \cite{sun2022efficient}, MotionSeg3D v2 (with the point refine module), and RVMOS \cite{kim2022rvmos}; b) an offline method: AutoMOS \cite{chen2022automatic}; c) point-based methods: 4DMOS (with the binary Bayes filter) \cite{mersch2022receding} and InsMOS \cite{wang2023insmos}; d) and another BEV-based method: LiMoSeg \cite{mohapatra2021limoseg}. To ensure a comprehensive comparison, we also provide all the evaluation results on the validation set (sequence 08). The experimental outcomes for each method are sourced from the SemanticKITTI-MOS benchmark and their respective original papers. Notably, all reported results are derived from training on the original SemanticKITTI dataset, and any results based on additional data \cite{sun2022efficient, chen2022automatic, wang2023insmos} are omitted to maintain fairness. 

As depicted in Table \ref{tab:table-benchmark}, RVMOS demonstrates superior performance on the hidden test set, which can be primarily attributed to its utilization of both semantic labels and moving object labels during training. Our method achieves the best result among approaches that solely relied on moving object labels, with an IoU of 69.7\%.

\begin{table}[h]
	\caption{Evaluation and Comparison on the SemanticKITTI Validation Set and the SemanticKITTI-MOS Benchmark. }
	\label{tab:table-benchmark}
	\centering
	\resizebox{0.8\linewidth}{!}{
		\begin{tabular}{lcc}
			\toprule
			{Methods} & IoU (Validation)  & IoU (Test) \\ 
			\midrule
			LMNet            &   66.4                  & 58.3                  \\
			MotionSeg3D, v1  &   68.1                   & 62.5                     \\
			MotionSeg3D, v2  &   71.4                   & 64.9                     \\
			AutoMOS          &   -                      & 54.3                     \\
			4DMOS            &   \textbf{77.2}          & 65.2                     \\
			LiMoSeg          &   52.6                   & -                    \\
			Ours, without delay   &   68.1                   & 63.9  \\
			Ours             &   76.5                   & \textbf{69.7}                     \\
			\midrule    
			LMNet$^{\ast}$            &   67.1                      & 62.5                  \\
			RVMOS$^{\ast}$            &   71.2                   & \textbf{74.7}      \\ 
			InsMOS$^{\star}$           &   73.2                   & 70.6       \\ 
			\bottomrule
		\end{tabular}
	}
	\begin{tablenotes}
		\footnotesize
		\item $^{\ast}$ indicates the method exploiting semantic labels.
		\item $^{\star}$ indicates the method exploiting instance labels.
		\item - indicates that the result is not available yet.
		\item Best results in bold.
	\end{tablenotes}
\end{table}

Fig. \ref{Fig: comparison} shows the qualitative comparison of LMNet, MotionSeg3D, 4DMOS, and our method on the SemanticKITTI validation set. Range view-based methods like LMNet are prone to boundary-blurring issues caused by back-projection, and MotionSeg3D partially improved this issue through the point refine module. However, due to the sensitivity of residual range image to distance, MotionSeg3D sometimes makes mistakes in classifying distant objects. 4DMOS performs well in segmenting dynamic objects at various distances but exhibits less distinct boundaries between objects. In contrast, our method effectively fuses appearance and motion features based on BEV representation and shows superior performance.

\subsection{Ablation Study}
In this section, we perform several ablation experiments to assess the contribution of our motion features and network components to the MOS performance. All experiments are conducted on the SemanticKITTI validation set (sequence 08).

As shown in Table \ref{tab:ablation}, we use vanilla PolarNet as the baseline, and vertically compare three different ways of fusing appearance and motion features: a) directly concatenating appearance and motion features (DC) following LMNet, b) exploring feature interactions with dual-branch structure bridged by motion guided attention module (MGA), and c) incorporating co-attention gate (CAG) on top of b) to learn appearance-motion co-attention. In addition, we also horizontally evaluate the performance of these network setups using different motion features as input for cross-comparison: range view-based motion features ($ F^m_{RV}$), and our bird's eye view-based motion features ($ F^m_{BEV}$). Both motion features are generated with consecutive 8 frames.

In general, we observe improvements in IoU for all setups using the proposed BEV-based motion features. When employing a dual-branch structure with motion guided attention module for deep multi-modality interaction, the setup with BEV-based motion features gets a stronger performance boost compared to the one with RV-based motion features. This indicates that our BEV-based motion features provide better temporal information compared to LMNet's range residual features. By combining the co-attention gate and motion-guided attention module, our method achieves the best performance and obtains a 16.6 percentage points improvement compared to the baseline. Overall, using BEV-based motion features and fusing appearance and motion features with the appearance-motion co-attention module (CAG + MGA) results in the best performance.

\begin{table}[t]
	\vspace{-10pt}
	\caption{Ablation Study of Network Components and Motion features on the SemanticKITTI Validation Set.}
	\label{tab:ablation}
	\resizebox{\linewidth}{!}{
		\begin{tabular}{l|cc|cc}
			\toprule
			\multirow{2}{*}{Baseline and components} & \multicolumn{2}{c|}{$ F^m_{RV}$} & \multicolumn{2}{c}{$ F^m_{BEV}$}    \\
			& IoU [\%]   & $ \Delta$ & IoU [\%] & $ \Delta$ \\ 
			\midrule
			PolarNet (without $F^m$)      & 59.99          & -         & 59.99               & -          \\
			PolarNet + DC   & 66.31         & +6.32     & 70.78               & +10.79          \\
			PolarNet + MGA       & 67.21    & +7.22     & 75.74               & +15.75          \\
			PolarNet + CAG + MGA & 69.40    & +9.41     & \textbf{76.54}      & \textbf{+16.55} \\ 
			\bottomrule
		\end{tabular}  
	}
	\begin{tablenotes}
		\footnotesize
		\item $\Delta$ means the improvement compared to the baseline.
	\end{tablenotes}
\end{table}
\begin{figure}[t]  
	\vspace{-12pt}
	\centering
	\includegraphics[width=1\linewidth]{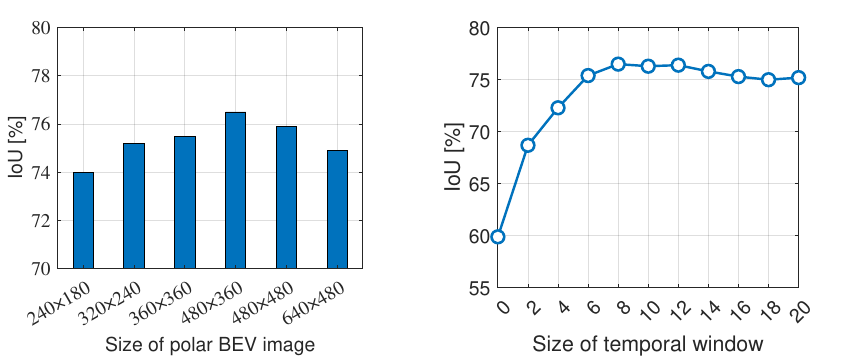}
	\vspace{-20pt}
	\caption{Ablation studies. The left figure shows the ablation study on the MOS performance vs. the size of the BEV image. The right figure shows the ablation study on the MOS performance vs. the temporal window size $N$.}
	\label{Fig: ablation}	
	\vspace{-15pt}
\end{figure}

Another ablation study is about the settings of motion features generation, as presented in Fig. \ref{Fig: ablation}. Firstly, we compare the MOS performance of using different BEV image sizes for motion features generation. All experiments are under the same settings except for the different representation sizes. The best-performing model employs polar BEV motion features with a representation size of 480$\times$360. Excessive enlargement of BEV images leads to the presence of numerous empty or sparsely populated grid cells, posing challenges in computing height differences and generating meaningful motion features. Furthermore, larger BEV image sizes render the motion features generation overly sensitive to abrupt or small objects in the environment, resulting in the emergence of unnecessary residuals.

We also examine the influence of the temporal window size $N$ on MOS performance. Since two sub-maps of the same size need to be constructed, $N$ must be set to an even number. It is noteworthy that the network achieves a high IOU even when $N$ is 0, indicating the absence of motion features. One possible reason is that the BEV approach effectively captures the spatial relationships between objects in the scene, enabling the network to infer the motion status of objects based on their positions on the road. We observe that the MOS performance improves as $N$ increases, reaching its peak in the range of 8 to 12. However, the further increase of $N$ leads to a slight decline in accuracy. This could be attributed to the temporal window becoming excessively long, causing the first sub-map $Q_1$ to be too distant from the current frame, which may result in erroneous motion features. This issue is particularly evident in scenes with significant field-of-view changes, such as during vehicle turning.

\begin{table}[t]
	\vspace{-10pt}
	\caption{Evaluation and Comparison on SipailouCampus Dataset.}
	\label{tab:livox}
	\centering
	\resizebox{0.9\linewidth}{!}{
		\begin{tabular}{lccc}
			\toprule
			Methods     & Retrain & IoU (Validation) & IoU (Test)     \\ 
			\midrule
			LMNet       &          & 5.37           & 6.88         \\
			MotionSeg3D &          & 6.83            &6.72          \\
			4DMOS       &          & \textbf{78.54}   & \textbf{82.30} \\
			Ours        &          & 50.44            & 52.02          \\
			Ours-h      &          & 70.94            & 71.51          \\
			\midrule
			LMNet       & \checkmark        & 54.27            & 56.16          \\
			MotionSeg3D & \checkmark        & 65.64            & 66.84          \\
			4DMOS       & \checkmark        & 87.30            & 88.89          \\ 
			Ours        & \checkmark        & \textbf{89.22}   & \textbf{90.80} \\ 
			\bottomrule
		\end{tabular}
	}
	\begin{tablenotes}
		\footnotesize
		\item Best results in bold.
	\end{tablenotes}
\end{table}
\begin{figure}[h]  
	\vspace{-7pt}
	\centering
	\includegraphics[width=1\linewidth]{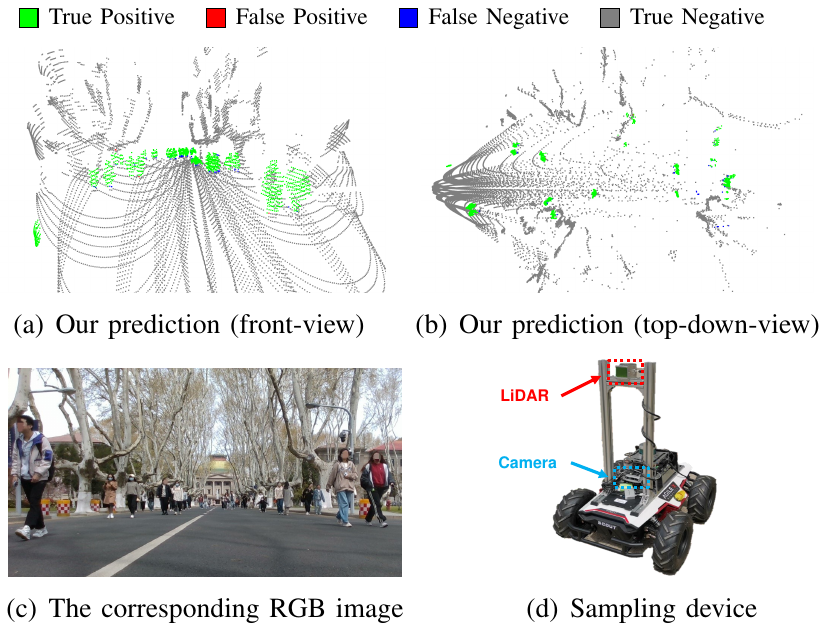}
	\vspace{-20pt}
	\caption{Visualization of MOS result of a high dynamic scene on SipailouCampus validation set.}
	\label{Fig: livox}	
	\vspace{-15pt}
\end{figure}

\subsection{Evaluation on SipailouCampus}

In this section, we demonstrate the performance of our method on a solid-state LiDAR with a small field of view and non-repetitive scanning mode. First, in order to test the generalization ability of the network, we exclude the intensity channel of the point cloud, train all the models on the training set of SemanticKITTI, and evaluate them on the validation and test set of SipailouCampus dataset. The results are shown in Table \ref{tab:livox}. Due to the significant differences in the range image projection between Velodyne LiDAR and Livox LiDAR, LMNet and MotionSeg3D (with the point refine module) fail to obtain reliable predictions. Our method does not perform satisfactorily without any parameter modification, as the BEV projection causes information loss in the vertical direction, making it difficult for the network to effectively segment objects within the same vertical grid (such as trees, pedestrians under trees, and the ground). By changing the overall height of the point cloud to match the ground position near the sensor with the SemanticKITTI dataset, we obtained a significant performance improvement without any additional training (corresponding to "Ours-h" in Table \ref{tab:livox}). In comparison, the method 4DMOS (with the binary Bayes filter), which processes directly on the point cloud, demonstrates strong generalization ability.

Furthermore, we retrain all models on the training set of SipailouCampus to obtain the final performance of all methods on Livox LiDAR data. To effectively capture and process LiDAR data within a different field of view, we make adjustments to the image size of projection-based methods. Specifically, the range image size of LMNet and MotionSeg3D is set to 512$\times$512, and the BEV image size of our method is set to 480$\times$64. Due to the irregular scanning pattern of Livox LiDAR, range view-based methods (LMNet, MotionSeg3D) still cannot achieve high accuracy. After retraining, our method outperforms 4DMOS, and reaches 89.3\% IoU on the validation set and 90.8\% IoU on the test set. Fig. \ref{Fig: livox} presents an example of the prediction results of our network.

\vspace{-5pt}
\subsection{Runtime}
We assess the runtime of different approaches by measuring the average inference time (ms) on the SemanticKITTI validation set, as shown in Table \ref{tab:table-runtime}. The evaluations for LMNet, MotionSeg3D, 4DMOS, and our method are conducted on a machine with an Intel Core i9-12900K CPU and a single NVIDIA RTX 3090 GPU. The results for RVMOS and InsMOS are obtained from their respective papers and are also evaluated using the RTX 3090 GPU. Our method achieves an inference time of only 23ms. However, incorporating the full set of motion features from all channels within the temporal window would lead to an extra fixed delay of $(N-1) \times 100$ ms. In practical usage, one can balance accuracy and runtime by controlling the length of the temporal window or deciding whether to use the complete motion features or not.
\begin{table}[h]
	\vspace{-10pt}
	\centering
	\caption{Comparison of Inference Time (ms) on SemanticKITTI Validation Set.}
	\label{tab:table-runtime}
	\resizebox{0.95\linewidth}{!}{ 
		\begin{tabular}{c|c|c|c|c|c}
			\toprule
			LMNet & MotionSeg3D & RVMOS & 4DMOS  &  InsMOS  & Ours \\
			\midrule
			35    & 110         & 29    & 132    &  127     & \textbf{23}   \\ 
			\bottomrule
		\end{tabular}
	}
\end{table}


\vspace{-5pt}
\section{Conclusions}

In this paper, we present a simple yet effective method for LiDAR-based online moving object segmentation. Our method exploits the spatio-temporal information from appearance and motion features in the bird's eye view domain and performs multi-modality feature fusion with a dual-branch network bridged by the appearance-motion co-attention module. Evaluation on the SemanticKITTI-MOS dataset demonstrates the proposed MotionBEV is the most accurate method among approaches utilizing only MOS labels, while also offering a balanced compromise between accuracy and latency. Moreover, experimental results on a dataset collected by a Livox LiDAR show our method's practical effectiveness on different types of LiDAR sensors. 





\renewcommand*{\bibfont}{\scriptsize}
\begin{spacing}{1}
	\scriptsize
    \printbibliography

@inproceedings{zhang2014loam,
  title={LOAM: Lidar odometry and mapping in real-time.},
  author={Zhang, Ji and Singh, Sanjiv},
  booktitle={Robot.: Sci. Syst.},
  volume={2},
  number={9},
  pages={1--9},
  year={2014},
  organization={Berkeley, CA}
}

@inproceedings{chen2019suma++,
	title={Suma++: Efficient lidar-based semantic slam},
	author={Chen, Xieyuanli and Milioto, Andres and Palazzolo, Emanuele and Giguere, Philippe and Behley, Jens and Stachniss, Cyrill},
	booktitle={Proc. IEEE/RSJ Int. Conf. Intell. Robots Syst.},
	pages={4530--4537},
	year={2019},
	organization={IEEE}
}

@article{guo2022obstacle,
	title={Obstacle avoidance with dynamic avoidance risk region for mobile robots in dynamic environments},
	author={Guo, Binghua and Guo, Nan and Cen, Zhisong},
	journal={IEEE Robot. Automat. Lett.},
	volume={7},
	number={3},
	pages={5850--5857},
	year={2022},
	publisher={IEEE}
}

@article{luo2018porca,
	title={Porca: Modeling and planning for autonomous driving among many pedestrians},
	author={Luo, Yuanfu and Cai, Panpan and Bera, Aniket and Hsu, David and Lee, Wee Sun and Manocha, Dinesh},
	journal={IEEE Robot. Automat. Lett.},
	volume={3},
	number={4},
	pages={3418--3425},
	year={2018},
	publisher={IEEE}
}

@inproceedings{kim2020remove,
	title={Remove, then revert: Static point cloud map construction using multiresolution range images},
	author={Kim, Giseop and Kim, Ayoung},
	booktitle={Proc. IEEE/RSJ Int. Conf. Intell. Robots Syst.},
	pages={10758--10765},
	year={2020},
	organization={IEEE}
}

@article{lim2021erasor,
	title={ERASOR: Egocentric ratio of pseudo occupancy-based dynamic object removal for static 3D point cloud map building},
	author={Lim, Hyungtae and Hwang, Sungwon and Myung, Hyun},
	journal={IEEE Robot. Automat. Lett.},
	volume={6},
	number={2},
	pages={2272--2279},
	year={2021},
	publisher={IEEE}
}

@article{chen2021moving,
	title={Moving object segmentation in 3D LiDAR data: A learning-based approach exploiting sequential data},
	author={Chen, Xieyuanli and Li, Shijie and Mersch, Benedikt and Wiesmann, Louis and Gall, J{\"u}rgen and Behley, Jens and Stachniss, Cyrill},
	journal={IEEE Robot. Automat. Lett.},
	volume={6},
	number={4},
	pages={6529--6536},
	year={2021},
	publisher={IEEE}
}

@inproceedings{sun2022efficient,
	title={Efficient Spatial-Temporal Information Fusion for LiDAR-Based 3D Moving Object Segmentation},
	author={Sun, Jiadai and Dai, Yuchao and Zhang, Xianjing and Xu, Jintao and Ai, Rui and Gu, Weihao and Chen, Xieyuanli},
	booktitle={Proc. IEEE/RSJ Int. Conf. Intell. Robots Syst.},
	pages={11456--11463},
	year={2022},
	organization={IEEE}
}

@article{kim2022rvmos,
	title={RVMOS: Range-View Moving Object Segmentation Leveraged by Semantic and Motion Features},
	author={Kim, Jaeyeul and Woo, Jungwan and Im, Sunghoon},
	journal={IEEE Robot. Automat. Lett.},
	volume={7},
	number={3},
	pages={8044--8051},
	year={2022},
	publisher={IEEE}
}

@article{mersch2022receding,
	title={Receding moving object segmentation in 3d lidar data using sparse 4d convolutions},
	author={Mersch, Benedikt and Chen, Xieyuanli and Vizzo, Ignacio and Nunes, Lucas and Behley, Jens and Stachniss, Cyrill},
	journal={IEEE Robot. Automat. Lett.},
	volume={7},
	number={3},
	pages={7503--7510},
	year={2022},
	publisher={IEEE}
}

@article{sun2020pointmoseg,
	title={PointMoSeg: Sparse tensor-based end-to-end moving-obstacle segmentation in 3-D lidar point clouds for autonomous driving},
	author={Sun, Yuxiang and Zuo, Weixun and Huang, Huaiyang and Cai, Peide and Liu, Ming},
	journal={IEEE Robot. Automat. Lett.},
	volume={6},
	number={2},
	pages={510--517},
	year={2020},
	publisher={IEEE}
}

@inproceedings{zhang2020polarnet,
	title={Polarnet: An improved grid representation for online lidar point clouds semantic segmentation},
	author={Zhang, Yang and Zhou, Zixiang and David, Philip and Yue, Xiangyu and Xi, Zerong and Gong, Boqing and Foroosh, Hassan},
	booktitle={Proc. IEEE Conf. Comput. Vis. Pattern Recognit.},
	pages={9601--9610},
	year={2020}
}

@inproceedings{yang2021learning,
	title={Learning motion-appearance co-attention for zero-shot video object segmentation},
	author={Yang, Shu and Zhang, Lu and Qi, Jinqing and Lu, Huchuan and Wang, Shuo and Zhang, Xiaoxing},
	booktitle={Proc. IEEE Int. Conf. Comput. Vis.},
	pages={1564--1573},
	year={2021}
}

@inproceedings{behley2019semantickitti,
	title={Semantickitti: A dataset for semantic scene understanding of lidar sequences},
	author={Behley, Jens and Garbade, Martin and Milioto, Andres and Quenzel, Jan and Behnke, Sven and Stachniss, Cyrill and Gall, Jurgen},
	booktitle={Proc. IEEE Int. Conf. Comput. Vis.},
	pages={9297--9307},
	year={2019}
}

@inproceedings{li2019motion,
	title={Motion guided attention for video salient object detection},
	author={Li, Haofeng and Chen, Guanqi and Li, Guanbin and Yu, Yizhou},
	booktitle={Proc. IEEE Int. Conf. Comput. Vis.},
	pages={7274--7283},
	year={2019}
}

@article{giraldo2020graph,
	title={Graph moving object segmentation},
	author={Giraldo, Jhony H and Javed, Sajid and Bouwmans, Thierry},
	journal={IEEE Trans. Pattern Anal. Mach. Intell.},
	volume={44},
	number={5},
	pages={2485--2503},
	year={2020},
	publisher={IEEE}
}

@inproceedings{zhao2022modeling,
	title={Modeling Motion with Multi-Modal Features for Text-Based Video Segmentation},
	author={Zhao, Wangbo and Wang, Kai and Chu, Xiangxiang and Xue, Fuzhao and Wang, Xinchao and You, Yang},
	booktitle={Proc. IEEE Conf.Comput. Vis. Pattern Recognit.},
	pages={11737--11746},
	year={2022}
}

@article{schauer2018peopleremover,
	title={The peopleremover-removing dynamic objects from 3-d point cloud data by traversing a voxel occupancy grid},
	author={Schauer, Johannes and N{\"u}chter, Andreas},
	journal={IEEE Robot. Automat. Lett.},
	volume={3},
	number={3},
	pages={1679--1686},
	year={2018},
	publisher={IEEE}
}

@inproceedings{pomerleau2014long,
	title={Long-term 3D map maintenance in dynamic environments},
	author={Pomerleau, Fran{\c{c}}ois and Kr{\"u}si, Philipp and Colas, Francis and Furgale, Paul and Siegwart, Roland},
	booktitle={Proc. IEEE Int. Conf. Robot. Automat.},
	pages={3712--3719},
	year={2014},
	organization={IEEE}
}

@inproceedings{fan2022dynamicfilter,
	title={DynamicFilter: an Online Dynamic Objects Removal Framework for Highly Dynamic Environments},
	author={Fan, Tingxiang and Shen, Bowen and Chen, Hua and Zhang, Wei and Pan, Jia},
	booktitle={Proc. IEEE Int. Conf. Robot. Automat.},
	pages={7988--7994},
	year={2022},
	organization={IEEE}
}

@article{chen2022automatic,
	title={Automatic labeling to generate training data for online LiDAR-based moving object segmentation},
	author={Chen, Xieyuanli and Mersch, Benedikt and Nunes, Lucas and Marcuzzi, Rodrigo and Vizzo, Ignacio and Behley, Jens and Stachniss, Cyrill},
	journal={IEEE Robot. Automat. Lett.},
	volume={7},
	number={3},
	pages={6107--6114},
	year={2022},
	publisher={IEEE}
}

@inproceedings{cortinhal2020salsanext,
	title={Salsanext: Fast, uncertainty-aware semantic segmentation of lidar point clouds},
	author={Cortinhal, Tiago and Tzelepis, George and Erdal Aksoy, Eren},
	booktitle={Proc. IEEE Veh. Symp. (IV)},
	pages={207--222},
	year={2020},
	organization={Springer}
}

@inproceedings{zhu2021cylindrical,
	title={Cylindrical and asymmetrical 3d convolution networks for lidar segmentation},
	author={Zhu, Xinge and Zhou, Hui and Wang, Tai and Hong, Fangzhou and Ma, Yuexin and Li, Wei and Li, Hongsheng and Lin, Dahua},
	booktitle={Proc. IEEE Conf. Comput. Vis. Pattern Recognit.},
	pages={9939--9948},
	year={2021}
}

@article{mohapatra2021limoseg,
	title={Limoseg: Real-time bird's eye view based lidar motion segmentation},
	author={Mohapatra, Sambit and Hodaei, Mona and Yogamani, Senthil and Milz, Stefan and Gotzig, Heinrich and Simon, Martin and Rashed, Hazem and Maeder, Patrick},
	journal={arXiv:2111.04875},
	year={2021}
}

@inproceedings{qi2017pointnet,
	title={Pointnet: Deep learning on point sets for 3d classification and segmentation},
	author={Qi, Charles R and Su, Hao and Mo, Kaichun and Guibas, Leonidas J},
	booktitle={Proc. IEEE Conf. Comput. Vis. Pattern Recognit.},
	pages={652--660},
	year={2017}
}

@inproceedings{behley2018efficient,
	title={Efficient Surfel-Based SLAM using 3D Laser Range Data in Urban Environments.},
	author={Behley, Jens and Stachniss, Cyrill},
	booktitle={Robot.: Sci. Syst.},
	volume={2018},
	pages={59},
	year={2018}
}

@inproceedings{berman2018lovasz,
	title={The lov{\'a}sz-softmax loss: A tractable surrogate for the optimization of the intersection-over-union measure in neural networks},
	author={Berman, Maxim and Triki, Amal Rannen and Blaschko, Matthew B},
	booktitle={Proc. IEEE Conf. Comput. Vis. Pattern Recognit.},
	pages={4413--4421},
	year={2018}
}

@inproceedings{geiger2012cvpr,
	author = {A. Geiger and P. Lenz and R. Urtasun},
	title = {{Are we ready for Autonomous Driving? The KITTI Vision Benchmark Suite}},
	booktitle = {Proc. IEEE Conf. Comput. Vis. Pattern Recognit.},
	pages = {3354--3361},
	year = {2012}
}

@article{xu2022fast,
	title={Fast-lio2: Fast direct lidar-inertial odometry},
	author={Xu, Wei and Cai, Yixi and He, Dongjiao and Lin, Jiarong and Zhang, Fu},
	journal={IEEE Trans. Robot.},
	volume={38},
	number={4},
	pages={2053--2073},
	year={2022},
	publisher={IEEE}
}

@article{everingham2010pascal,
	title={The pascal visual object classes (voc) challenge},
	author={Everingham, Mark and Van Gool, Luc and Williams, Christopher KI and Winn, John and Zisserman, Andrew},
	journal={Int. J. Comput. Vision},
	volume={88},
	pages={303--338},
	year={2010},
	publisher={Springer}
}

@inproceedings{hu2020randla,
	title={Randla-net: Efficient semantic segmentation of large-scale point clouds},
	author={Hu, Qingyong and Yang, Bo and Xie, Linhai and Rosa, Stefano and Guo, Yulan and Wang, Zhihua and Trigoni, Niki and Markham, Andrew},
	booktitle={Proc. IEEE Conf. Comput. Vis. Pattern Recognit.},
	pages={11108--11117},
	year={2020}
}

@article{ding2022self,
	title={Self-Supervised Scene Flow Estimation With 4-D Automotive Radar},
	author={Ding, Fangqiang and Pan, Zhijun and Deng, Yimin and Deng, Jianning and Lu, Chris Xiaoxuan},
	journal={IEEE Robot. Automat. Lett.},
	volume={7},
	number={3},
	pages={8233--8240},
	year={2022},
	publisher={IEEE}
}

@inproceedings{li2022rigidflow,
	title={Rigidflow: Self-supervised scene flow learning on point clouds by local rigidity prior},
	author={Li, Ruibo and Zhang, Chi and Lin, Guosheng and Wang, Zhe and Shen, Chunhua},
	booktitle={Proc. IEEE Conf. Comput. Vis. Pattern Recognit.},
	pages={16959--16968},
	year={2022}
}

@inproceedings{dong2022exploiting,
	title={Exploiting rigidity constraints for lidar scene flow estimation},
	author={Dong, Guanting and Zhang, Yueyi and Li, Hanlin and Sun, Xiaoyan and Xiong, Zhiwei},
	booktitle={Proc. IEEE Conf. Comput. Vis. Pattern Recognit.},
	pages={12776--12785},
	year={2022}
}

@inproceedings{kreutz2023unsupervised,
	title={Unsupervised 4D LiDAR Moving Object Segmentation in Stationary Settings with Multivariate Occupancy Time Series},
	author={Kreutz, Thomas and M{\"u}hlh{\"a}user, Max and Guinea, Alejandro Sanchez},
	booktitle={Proc. IEEE Winter Conf. Appl. Comput. Vis.},
	pages={1644--1653},
	year={2023}
}

@article{wang2023insmos,
	title={InsMOS: Instance-Aware Moving Object Segmentation in LiDAR Data},
	author={Wang, Neng and Shi, Chenghao and Guo, Ruibin and Lu, Huimin and Zheng, Zhiqiang and Chen, Xieyuanli},
	journal={arXiv preprint arXiv:2303.03909},
	year={2023}
}
\end{spacing}

\end{document}